\documentclass[runningheads]{llncs}
\usepackage{amsmath}
\usepackage{graphicx}
\usepackage{subfig}
\usepackage{authblk}

\begin{document}

\title{Technical Report of Mobile Manipulator Robot for Industrial Environments}
%
%\titlerunning{Abbreviated paper title}
% If the paper title is too long for the running head, you can set
% an abbreviated paper title here
%
\author{Erfan Amoozad Khalili\inst{1} \and
Kiarash Ghasemzadeh \inst{1} \and
Hossein Gohari\inst{1} \and 
Mohammadreza Jafari\inst{1} \and
Matin Jamshidi \inst{1} \and
Mahdi Khaksar \inst{1} \and
AmirReza AkramiFard \inst{2} \and
Mana Hatamzadeh\inst{2} \and 
Saba Sadeghi\inst{2} \and
Mohammad Hossein Moaiyeri \inst{3}
}
\authorrunning{E. Amoozad Khalili, M. Jamshidi, K. Ghasemzadeh , M. Jafari, et al.}
% First names are abbreviated in the running head.
% If there are more than two authors, 'et al.' is used.
%
\institute{Faculty of Electrical Engineering ,Shahid Beheshti University \\
\email{h\_moaiyeri@sbu.ac.ir}\\
\email{\{e.amouzadkhalili, k.ghasemzadeh, ho.gohari, mohammadr.jafari, mat.jamshidi, m.khaxar, a.akramifard, m.hatamzadeh, sab.sadeghi\}@mail.sbu.ac.ir}\\
\url{https://sbu-team.github.io/}}
\maketitle              % typeset the header of the contribution
\begin{abstract}
This paper describes Auriga’s @Work team and their robot, developed at Shahid Beheshti University Faculty of Electrical Engineering's Robotics and Intelligent Automation Lab for RoboCup 2024 competitions. The robot is designed for industrial tasks, optimizing efficiency in repetitive or hazardous environments. It features a 4-wheel Mecanum system for omnidirectional movement and a 5-degree-of-freedom manipulator arm with a 3D-printed gripper for object handling and navigation. The electronics include custom boards with ESP32 microcontrollers and an Nvidia Jetson Nano for real-time control. Key software components include Hector SLAM for mapping, A* path planning, and YOLO for object detection, supported by integrated sensors for enhanced navigation and collision avoidance.

\keywords{Robocup@Work \and Robotics  \and Robots Grasping \and Computer Vision \and }
\end{abstract}
\begin{figure}
\centering
\includegraphics[width=8cm]{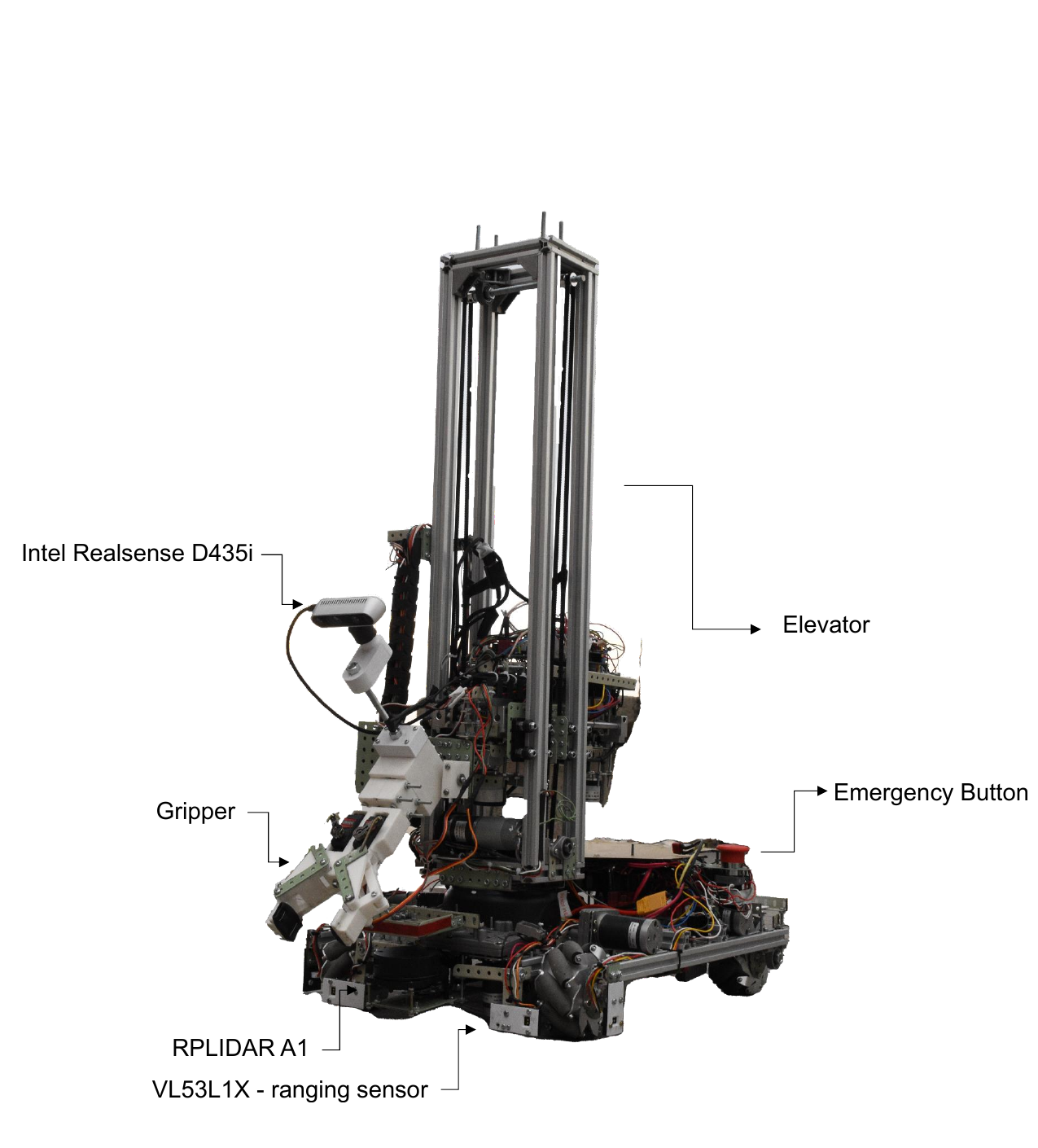}
\caption{photo of the robot.} \label{fig0}
\end{figure}
\section{Introduction}
Nowadays in many industries, robots are a suitable option for optimizing tasks and increasing efficiency. @Work robots can meticulously do tasks deemed repetitive, exhausting, impossible or dangerous for humans. On the other hand, robots can help people in the work environment, A combination of workers and robots can multiply efficiency. 
@Work robots are useful in a wide variety of industries. Contributing to a faster production line, assisting humans, limited environment transportation, doing tasks considered difficult for humans and many other desired tasks we can define for the robot are but a few of @Work robots’ applications.
\section{Hardware Description}
\subsection{Mechanical Design} 
\subsubsection{Body}
This part includes 4 Mecanum wheels and each wheel has a separate motor. Previously mentioned motors are DC motors, each one connected to the corresponding wheel by a solar gearbox. Because of using Mecanum wheels and separate motors each one of the wheels can move independently so it increases maneuverability. Robot’s chassis is built using Aluminum profiles.
\begin{figure}
\centering
\includegraphics[width=8cm]{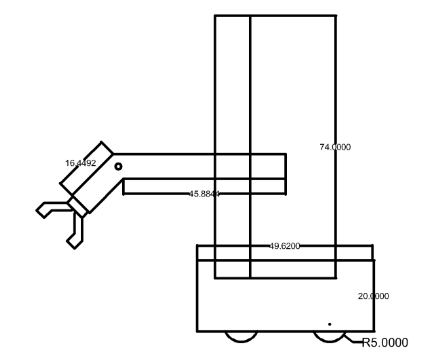}
\caption{Illustration of the robot body showing mechanical design.} \label{fig1}
\end{figure}

\subsubsection{Arm}
The robot’s arm has four sections
\begin{enumerate}

\item {\textbf{Rotary section}}\\
The rotary section consists of a set of ball bearings and surfaces that the arm is mounted on and turning this surface by a DC motor with gearboxes makes the arm rotate around the Z axis.\\
\item {\textbf{Elevator section}}\\
The elevator section consists of a set of 2 by 2 profiles, ball bearings, strap, pulley and etc. which moves the gripper back and forth by two DC Motors with gearboxes. (linear movement in X-axis)\\
\item {\textbf{Telescopic section}}\\
The telescopic section of the assembly consists of a set of guide shafts, linear bearings, belts, pulleys, etc. These components enable the gripper to move forward and backward in the direction of the X-axis using two DC-geared motors, providing linear motion.\\
\item {\textbf{Gripper}} \\
The mechanical structure of the gripper, depicted in Fig. 3, was entirely designed in Solid-Works and printed by a 3D printer to be able to grab different objects with different orientations. The gripper contains four Servo motors, two of which are used to grab objects and the other two are used to change the location and orientation of the end-effector.\\
\begin{figure}
\centering
\includegraphics[width=9cm]{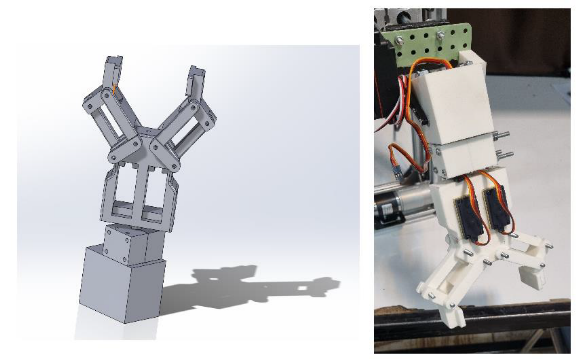}
\caption{3D model and 3D printed gripper used in robot's arm.} \label{fig2}
\end{figure}
\end{enumerate}
The body structure of the robot is as shown in Fig. 1. All mechanical sections are designed, built, and assembled entirely by the mechanics team and no pre-built parts have been used.

\subsection{Electrical Design} 
{Auriga’s @Work consists of two electronic boards, one board used in the top section of the robot to control the movement of robot’s arm and one board in the bottom section of the robot to control general movement of the robot which are connected by UART protocol. These boards are powered by a four cell 16000 mAH lithium polymer battery. Voltage regulators have been used in order to supply the required voltage of different circuits.}
\begin{enumerate}

\item{ \textbf{Arm’s board}}\\
In this board two XL4016 switching regulators supply servo motors, encoders and one AMS1117 supplies 3.3V for ESP32 microcontroller and ICs. This board contains one ESP32 microcontroller, two triple 2:1 MUX/DeMUX ICs, and two L298N drivers to control the movement of DC motors in the arm section of the robot. Also, The arm section contains four servo motors which are controlled by this board.
\begin{figure}
\centering
\includegraphics[width=6cm]{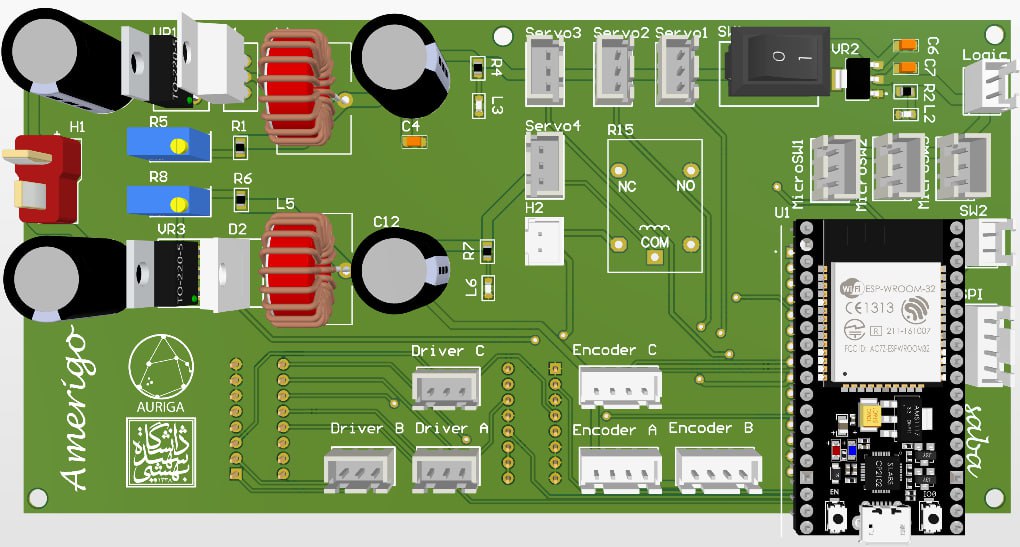}
\caption{Printed Circuit Board of arm's board.} \label{fig5}
\end{figure}
\item{ \textbf{Board located in the body}}\\
In this board, two XL4016 switching regulators supply encoders and Nvidia Jetson Nano board(5V), and one linear regulator AMS1117 supplies 3.3V for two microcontrollers, ICs, range measurement sensors, and gyroscopes. Gyroscopes and range measurement sensors send data to the desired microcontroller by I2C protocol. Also, microcontrollers and Jetson Nano’s connection and data exchange are done by Serial connection using UART protocol. Similar to the top section board, the designed circuit for controlling DC motors contains an L298N driver and a 2:1 MUX/DeMUX IC. Since we have 5 motors in this section, we have used three drivers.
It is possible to turn off the motors and turn them back on by switching the emergency switch key available on both boards. The designed board for this case contains one or multiple MUX ICs and relays.
All parts of designing the robot’s boards, designing PCBs using the Altium Designer Application, and assembling the parts have been done by the electronics team.

\begin{figure}
\centering
\includegraphics[width=6cm]{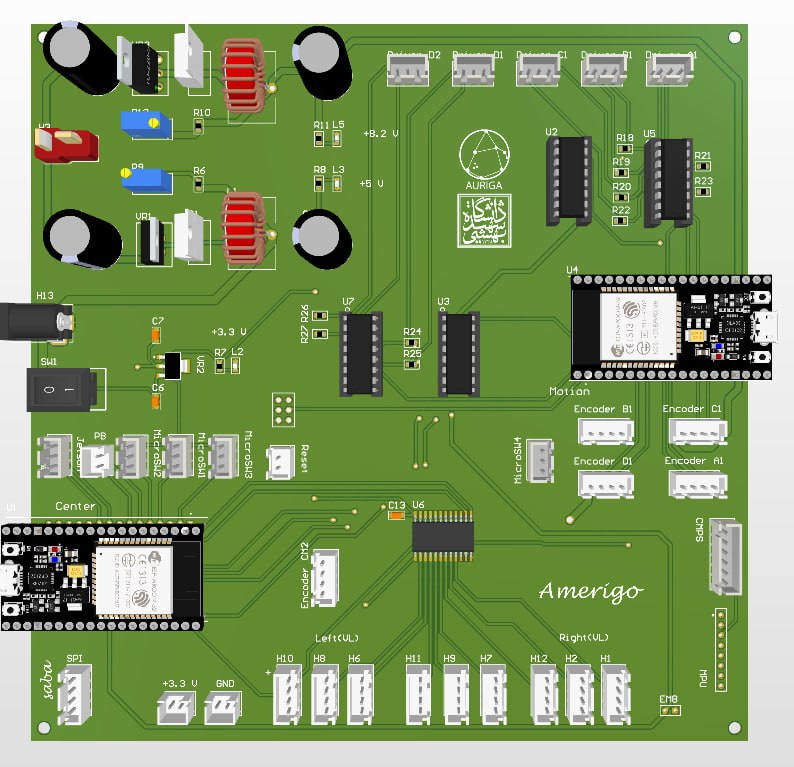}
\caption{Printed Circuit Board of the main board.} \label{fig4}
\end{figure}
\end{enumerate}
\subsection{Sensors}
\begin{enumerate}
    \item {\textbf{Intel RealSense D435i Depth Camera}}\\
    This sensor is a camera with the ability to detect depth and send RGB-D data to the main processor of the robot\cite{keselman2017intel}.
    \item {\textbf{RPLIDAR A1}}\\
    This time-of-flight range measurement sensor uses lasers to scan 360 degrees around itself and detect obstacles around the robot. This sensor is located in front of the robot and is used in implementing mapping and navigation algorithms. 
    \item {\textbf{Rotary Encoder}}\\
    We have used the E38S6G5-100B-G24N rotary encoder. This sensor calculates the movement of the desired section of the arm or angular velocity of wheels using the amount of motor’s shafts’ rotation. This sensor has been used both in the movement section of the body and in the arm. These sensors have been used to estimate arm position, and wheel velocity and generally for state estimation of different sections.
    \item {\textbf{Range Sensor}}\\
    These sensors are used in order to understand the distance of obstacles with respect to the robot for LiDAR’s blind spots. Said sensor is a time-of-flight range sensor VL53L1X which prevents collision when moving left, right, or back. The reason for using this sensor is its high speed and accuracy characteristics. 
    \item{\textbf{Gyroscope}}\\
    To perform angle measurements of the robot, CMPS14 and MPU6050 sensors have been used. After combining the incoming data of the two sensors, one final result will be used as the estimated angle. This sensor is also used for increasing the accuracy of the robot’s localization. 
\end{enumerate}

\section{Algorithm and Software full description}

\subsection{Localization and Mapping}
@Work robot needs an accurate map of the environment for searching the environment. In this regard, the robot obtains its estimated location with respect to a global coordinate system by combining the system’s motion model equations\cite{author2024kinematic}, encoders, and gyroscopes’ data. On the other hand, the robot obtains obstacles’ location in the surrounding environment in its local coordinate system by constantly knowing the location of itself and receiving LiDAR and auxiliary range measurement sensors’ data then locates obstacles present in the environment in the global coordinates using a transform function. Finally, a constructed map of the environment is saved in the processor’s memory as a two-dimensional array and will be used in the navigation algorithm.
\begin{figure}
\centering
\includegraphics[width=9cm]{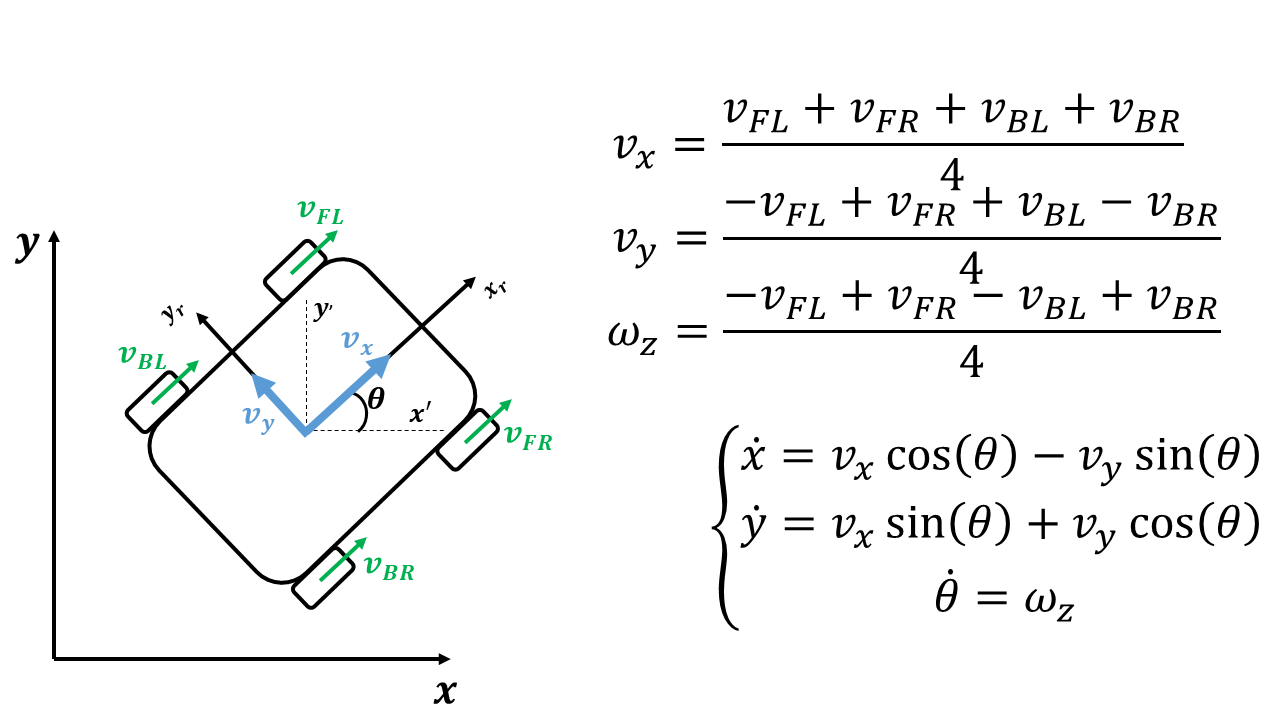}
\caption{Motion Model Equations for 4-Wheel Mecanum Robot} \label{fig6}
\end{figure}
We utilize the Hector SLAM algorithm for mapping and navigation within the workspace. In Figure \ref{fig:example}, you can see an example output of the Hector SLAM mapping \cite{KohlbrecherMeyerStrykKlingaufFlexibleSlamSystem2011}. The purple color represents the navigated path, the blue area represents the walls, the white area represents the free space, and the gray area represents the unmapped space.
\begin{figure}%
    \centering
    \subfloat[\centering Hector SLAM map]{{\includegraphics[width=4cm]{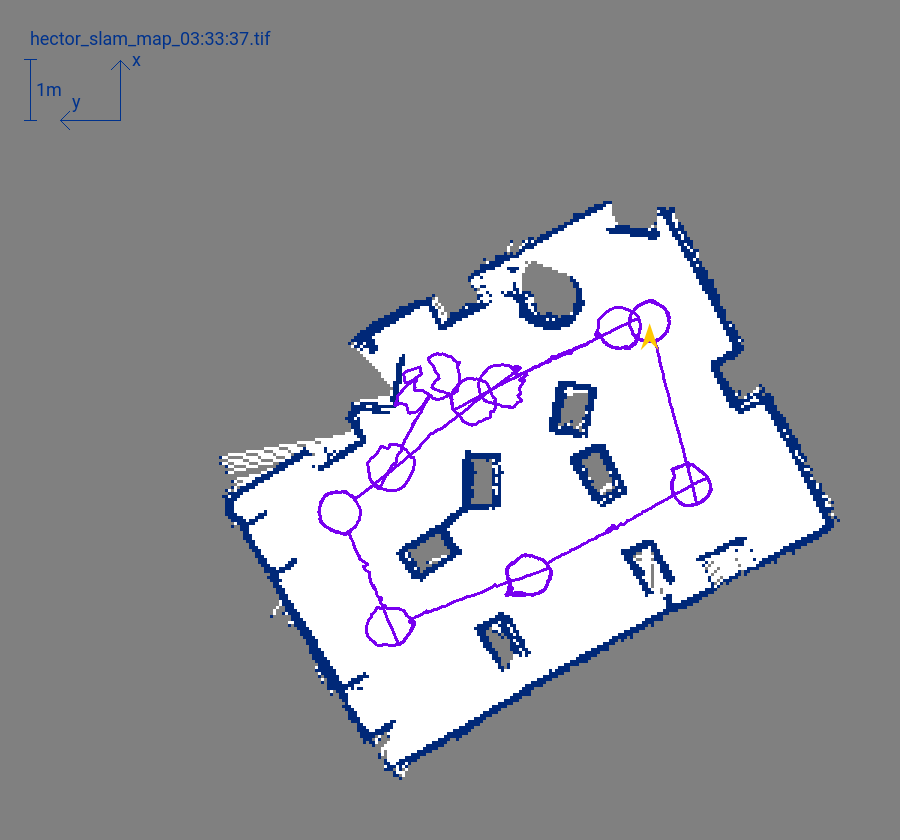} }}%
    \qquad
    \subfloat[\centering The mapped area]{{\includegraphics[width=7cm]{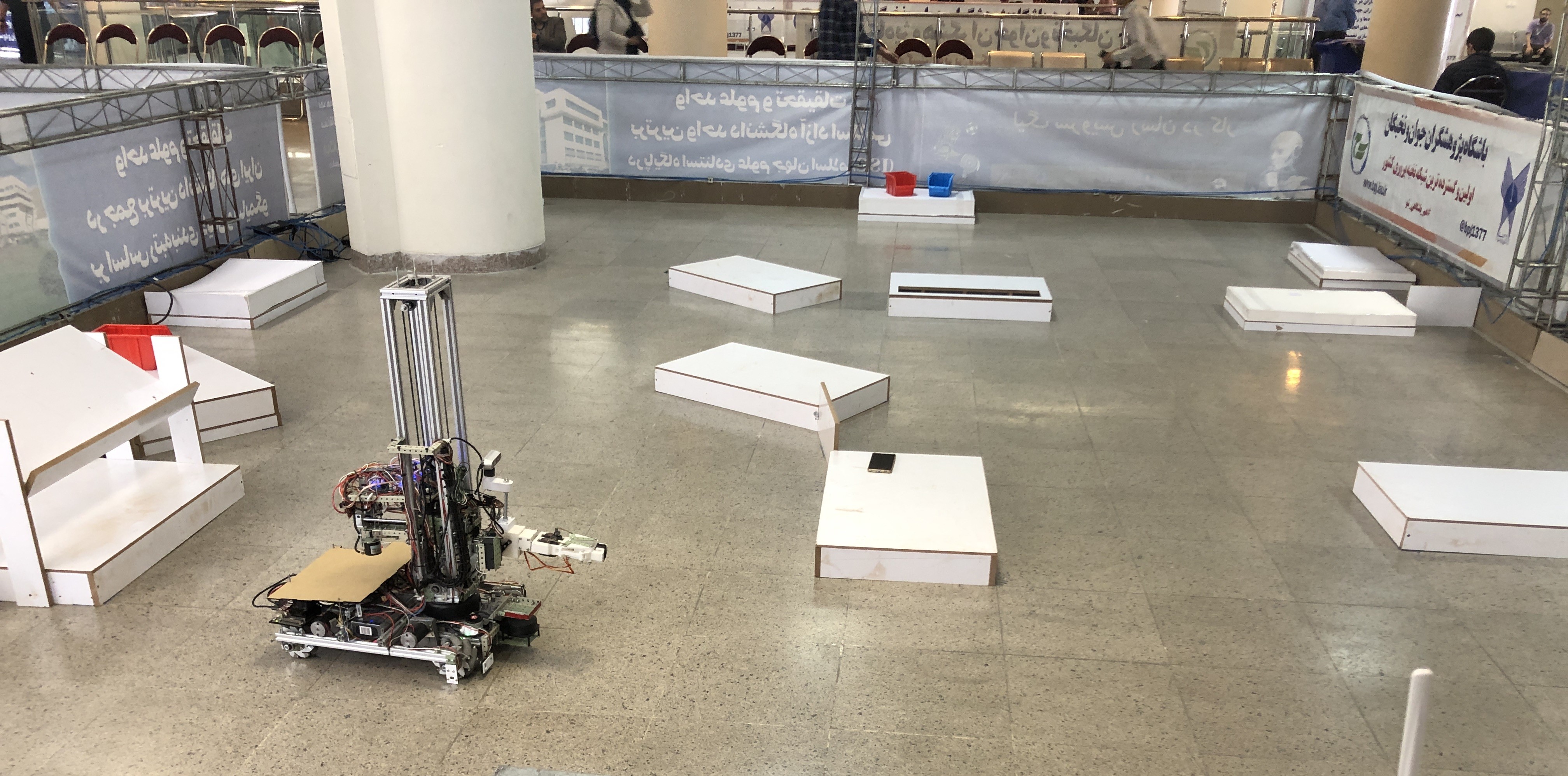} }}%
    \caption{SLAM algorithm}%
    \label{fig:example}%
\end{figure}
\subsection{Decision-Making and Navigation}
In this robot, we divide the decision-making process for doing the desired tasks into two parts:

\begin{enumerate}
\item {\textbf{Mission Planning}}
In the mission planning phase, the robot optimizes time and resources by selecting actions based on the current and desired state of the service area. As shown in the flowchart in Figure \ref{figflowchart}, the robot first assesses if the area is in the desired state. If not, it determines whether to fetch a new item or deliver a previously fetched one, based on a cost evaluation of both options. The goal is to minimize resource use, ensuring efficient task execution. Once a decision is made, the result is passed to the path planning phase, where the robot navigates to perform the chosen task.
\begin{figure}[h]
\centering
\includegraphics[width=0.9\textwidth]{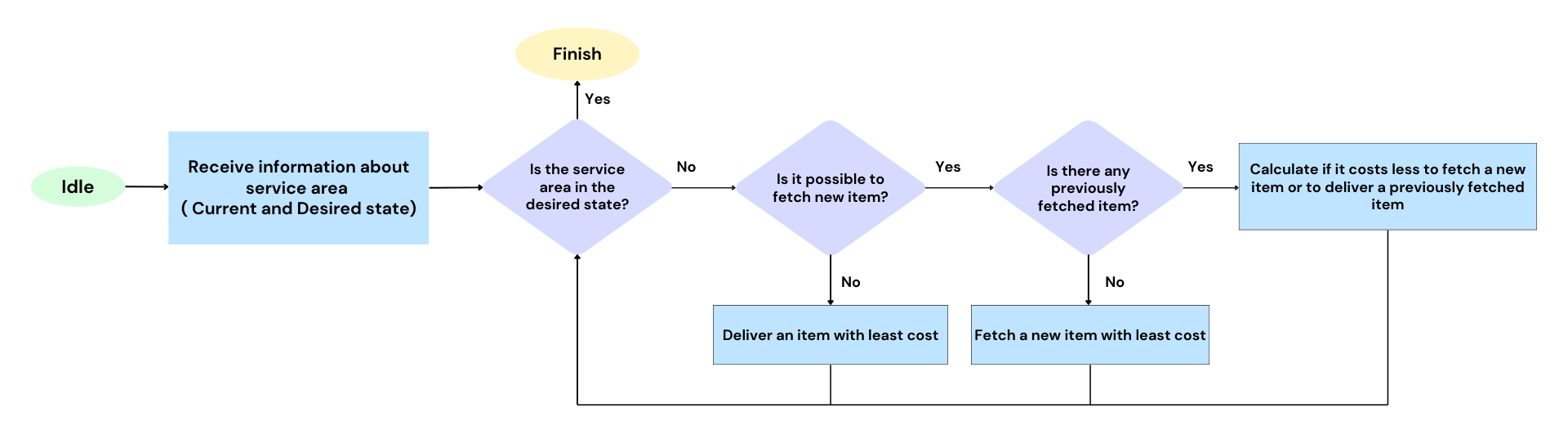}
\caption{Flowchart of Mission Planning Algorithms.} \label{figflowchart}
\end{figure}
\newpage
\item {\textbf{Path Planning}}\\
 Path planning is designed to make the robot move from the current location to the desired destination which is specified as the result of the mission planning algorithm, in path planning we have used A* \cite{Hart1968} \cite{6968200} algorithm to find the shortest route for the robot to go without colliding with the obstacles based on the map we have of the environment. The implementation of this algorithm utilizes a MinHeap data structure, which significantly affects the algorithm's execution time. It is important to note that the greater the distance from the start to the destination, the more pronounced this time difference becomes. This is because the order of searching among the open nodes changes from \(\mathcal{O}(n)\) to \(\mathcal{O}(\log n)\).
The cost function in the implementation of the A* algorithm is defined as the sum of G-cost, H-cost, and Safety-cost. The reason for adding the last term to the cost function is to increase the cost of cells that are closer to the walls, so if the robot needs to pass between two walls, This encourages the robot to move through the middle of the path, reducing the likelihood of colliding with the walls. It is also important to note that the Euclidean distance from the target cell to the destination cell is chosen as the criterion for H-cost.
\[
J = G + \sqrt{(x - x_{\text{final}})^2 + (y - y_{\text{final}})^2} + \text{Safety\,cost}
\]
\[
\text{Safety\,cost} = \frac{K}{\text{distance \,\, to \,\, the \,\, nearest \,\, wall}}
\]
After the path robot has to go is decided, that path is given to the control system as a reference, and the robot uses PID controller \cite{1453566} to remain in its path until reaching the destination. While moving towards the destination, LiDAR and range measurement sensors’ data is also checked in real-time and if there’s a possibility of collision in the path the robot stops (because of possible error in calculating the optimized path) and runs navigation process again.
For the implementation of said algorithms C++ programming language is used and these algorithms are processed real-time on the main processor, Nvidia Jetson Nano.\\

\end{enumerate}

\subsection{Computer Vision}
    Our computer vision system leverages the Intel RealSense D435i camera to achieve precise and efficient environmental perception, which is crucial for tasks such as object recognition, spatial mapping, and virtual boundary detection. The section is organized into three critical components that contribute to the system’s robust performance:

    \begin{enumerate}

    \item {\textbf{Object Segmentation:}}\\
    Objects located within each workstation are identified and segmented in real-time using the YOLO (You Only Look Once) neural network, a state-of-the-art object detection algorithm known for its high speed and accuracy in detecting multiple classes of objects in a single forward pass \cite{Jocher_Ultralytics_YOLO_2023} \cite{7780460}. The YOLO model allows the system to recognize diverse objects, such as tools, products, or obstacles, facilitating efficient task automation and enhanced decision-making processes. Once detected, the objects are segmented from the background, enabling further processing and manipulation. Example of object detection is shown in Figure \ref{figfff}.

    \item {\textbf{Locating Objects:}}\\
    Once objects are segmented and classified by the YOLO neural network, the system utilizes the depth data provided by the Intel RealSense D435i camera to compute the precise spatial location of each object. The camera's depth sensing capabilities allow for accurate extraction of x, y, and z coordinates, providing a 3D understanding of the environment.
    For angle detection we use the PCA algorithm \cite{pca}. This spatial information is essential for applications such as robotic grasping, navigation, or further object interaction, as it enables the system to determine the object’s distance, orientation, and position relative to the camera and other surrounding objects. 
    \begin{figure}
        \centering
        \includegraphics[width=0.45\linewidth]{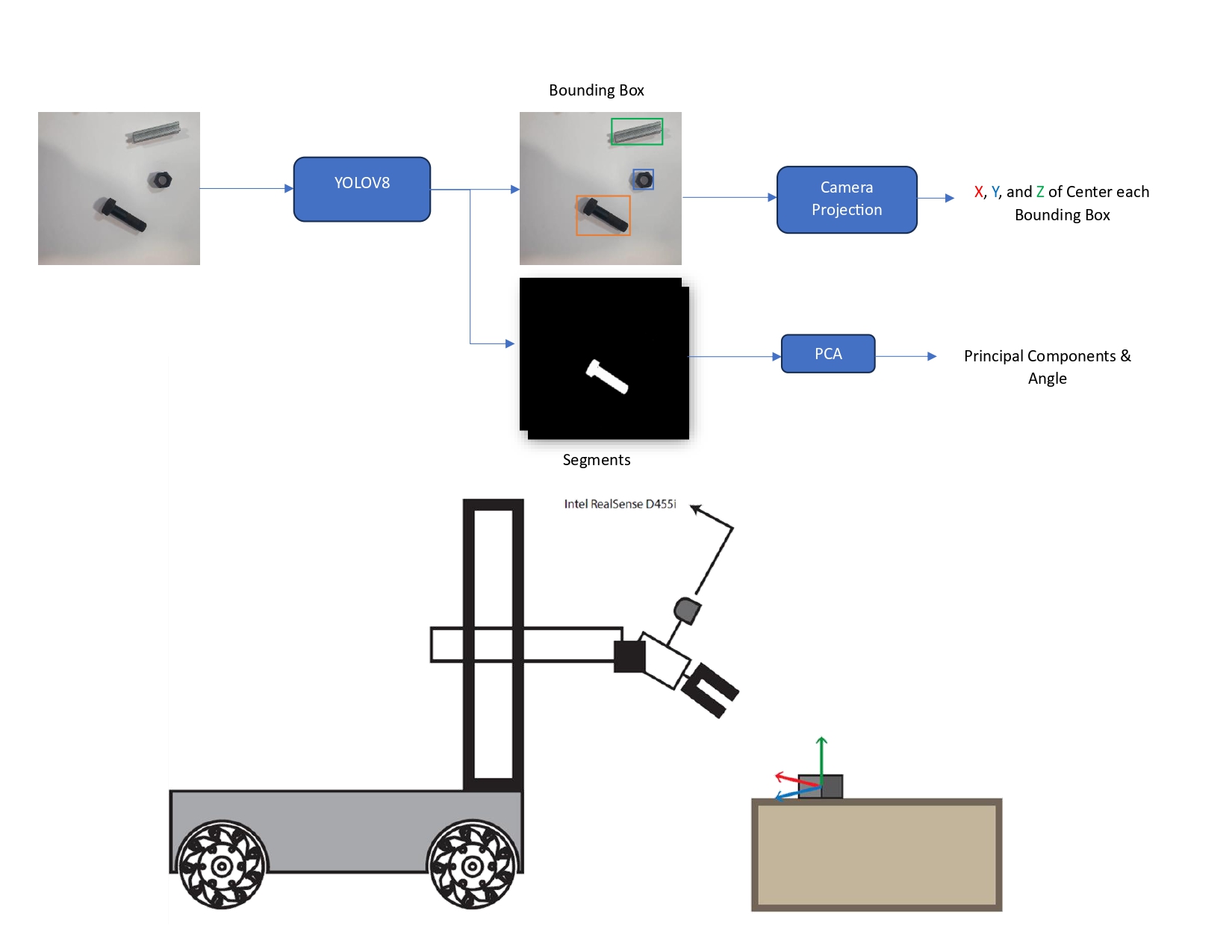}
        \caption{Illustration of Object Pose Estimation Algorithm.}
        \label{fig:enter-label}
    \end{figure}
\begin{figure}[b] 
        
        \centering
        \subfloat[\centering Example of Object Detection using YOLO]{{\includegraphics[width=3.5cm]{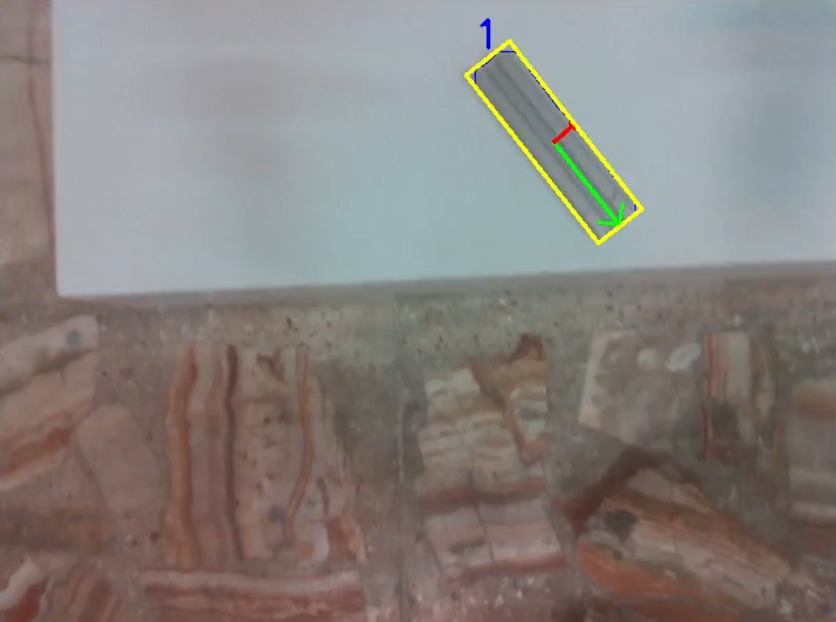} }}%
        \qquad
        \subfloat[\centering Example of Detection of Multiple Objects]{{\includegraphics[width=3.5cm]{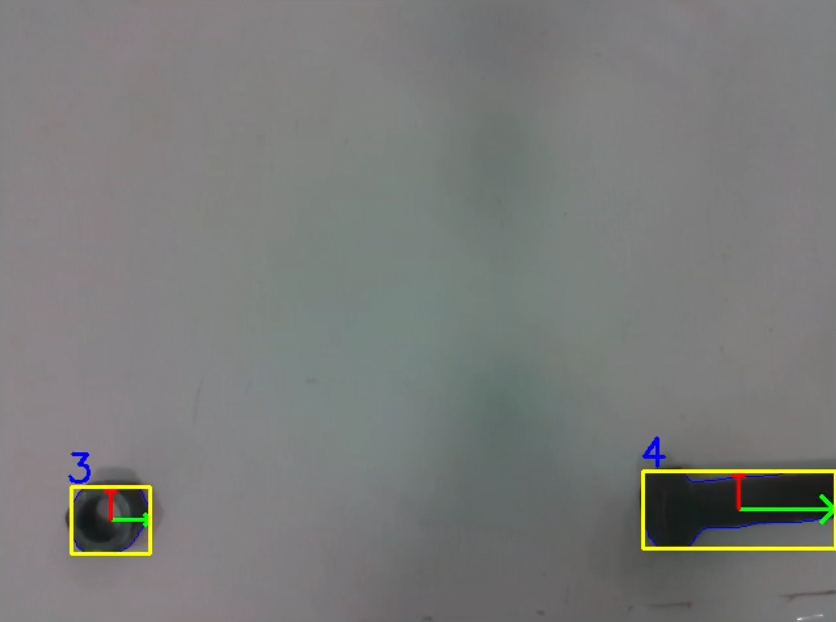} }}%
        \caption{Examples of Object Detection and Segmentation in Various Stations}%
    \label{figfff} 
    \end{figure}
    \newpage
    \item {\textbf{Virtual Walls Detection:}}\\
    Virtual walls or barriers are detected through the analysis of point clouds generated by the depth camera. The point cloud data, consisting of a dense collection of 3D points, is used to model the surrounding environment, enabling the system to identify and locate virtual walls or restricted zones. These virtual walls can be critical for defining boundaries for robotic movement or ensuring safety by preventing collision with off-limits areas. An example of virtual walls is mentioned in Figure \ref{fig55}.\\
\begin{figure}[h]
\centering
\includegraphics[width=5cm]{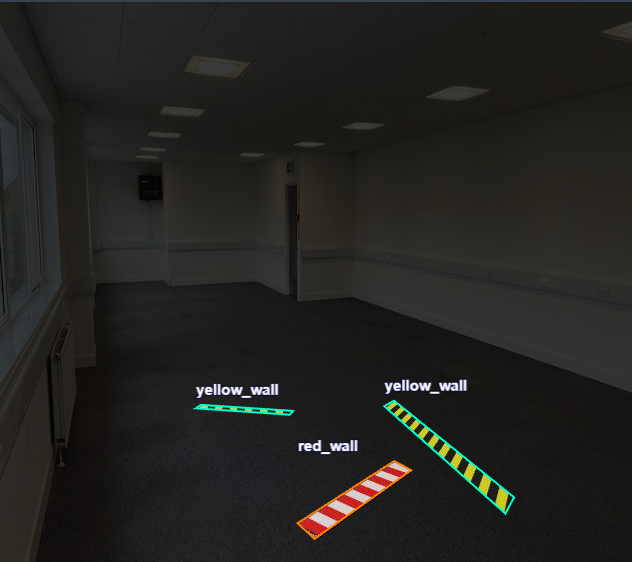}
\caption{Virtual Walls Annotations.} \label{fig55}
\end{figure}
\end{enumerate}

\subsection{Pick and Place Objects}
\begin{figure}[t]
\centering
\includegraphics[width=8cm]{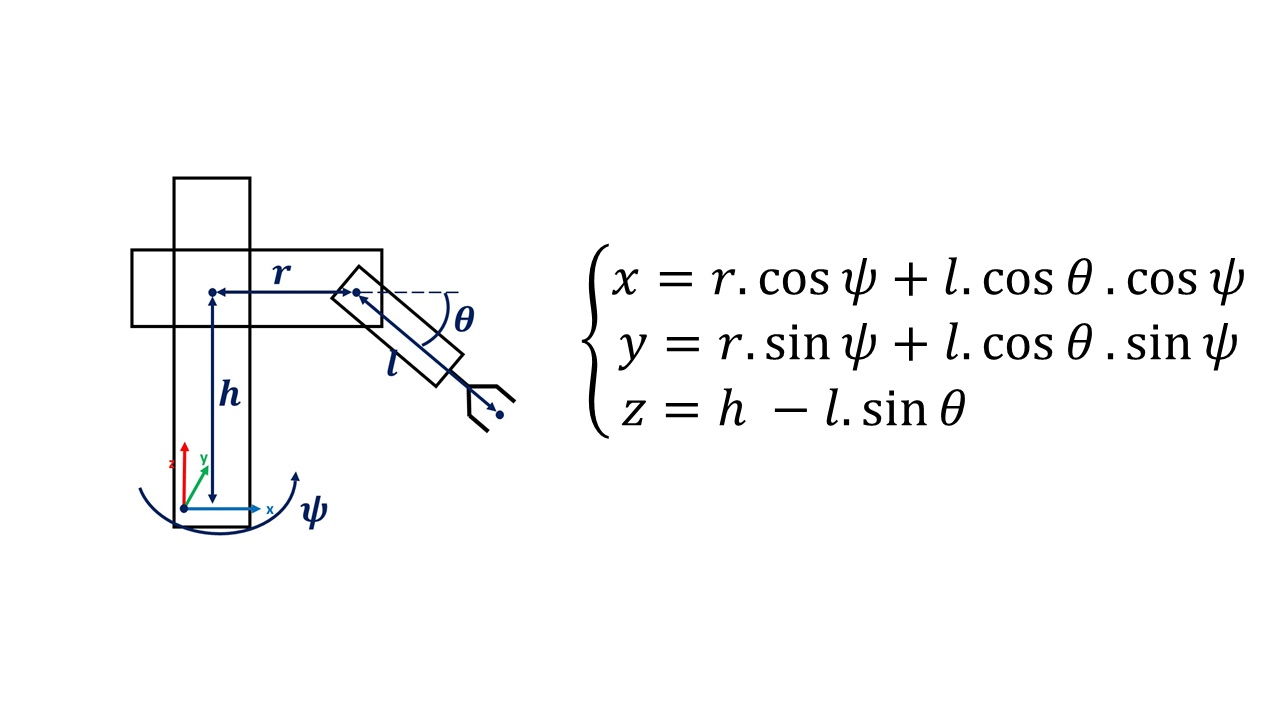}
\caption{Arm Kinematics} \label{fig66}
\end{figure}

The manipulator is a 5-degree-of-freedom arm, with each motor in a closed-loop system using sensors to ensure accurate real-time state estimation. A finite-state machine has been used for grasping objects. To pick up an object, the arm first uses feedback from the camera data to align its y-axis position with the target object. Then, it adjusts the telescopic part of the arm along the x-axis to align with the object. Finally, it matches its angle and height with the object and closes the gripper to grab it. The equations for the position of the arm's end-effector based on the status of each motor is mentioned in Figure \ref{fig66}
\newpage
\subsection{Robot Operating System}
This section is divided into two important parts. The first part is receiving instructions from AC \cite{Kraetzschmar2014} and sending these instructions to the relevant sections in the robot and in the second part AI and computer vision, receiving data from the camera and LiDAR, mapping, navigation, and control algorithms are placed next to in each other in the ROS platform. At times when a process is useless, that process is shut down to use the resources in the best way and in an optimal way\cite{ros}.

\subsection{Embedded System Design}
\begin{figure}
\includegraphics[width=\textwidth]{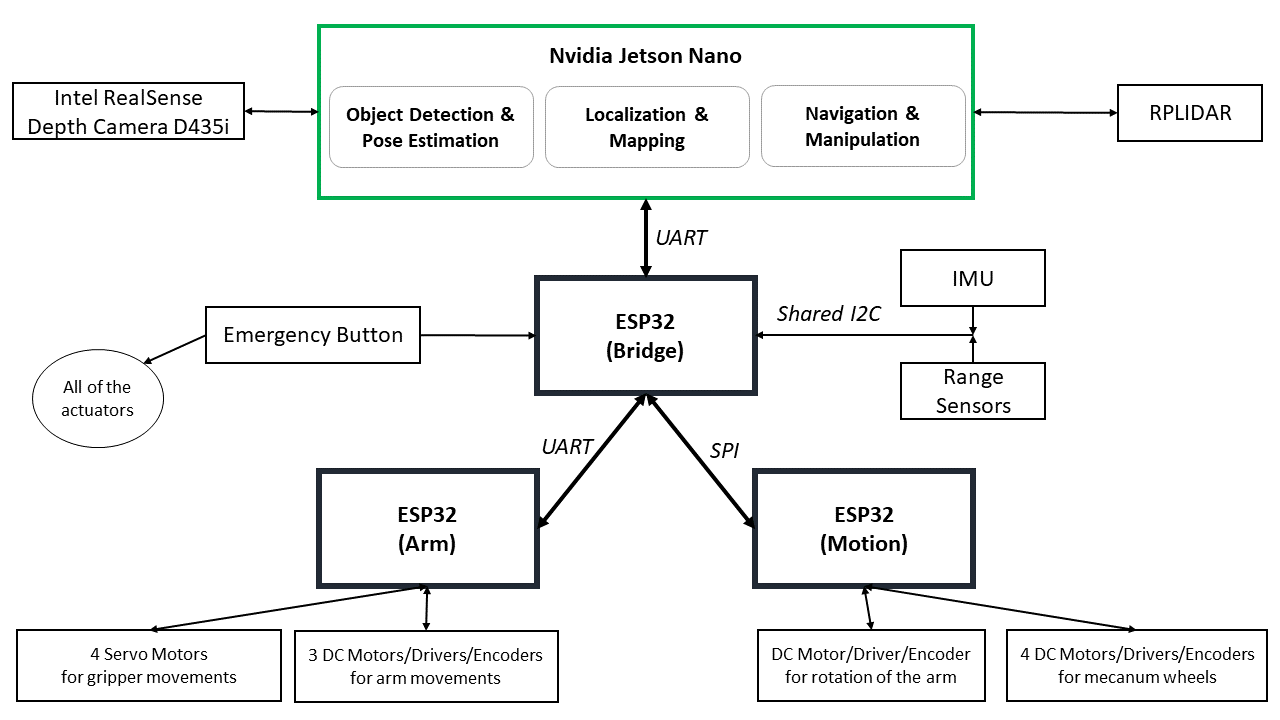}
\caption{Embedded System Architecture.} \label{fig3}
\end{figure}

In order to control, and drive body and arm’s motors, receive data from encoders, gyroscopes, range measurement sensors and communication between processors an embedded system has been implemented which contains three ESP32 microcontrollers, sensors, and appropriate drivers. Required processes are done in Nvidia Jetson Nano board and the final command is sent to the microcontroller that must execute that command and the destination microcontroller sets the speed and location of the motor internally. We are using incremental PID controller \cite{Li_2023} to precisely adjust mecanum wheels speed based on encoders feedback and standard PID controller to control the position, ensuring smooth and accurate movement across all components. The connection between ESP32s is built by an SPI protocol with a common Bus and Jetson Nano and the main ESP32’s connection is through UART protocol. The hardware system’s general schematic is shown in Figure \ref{fig3}.
\begin{table}
\centering
\caption{General Characteristics}\label{tab1}
\begin{tabular}{|l|l|}
\hline
Attrbiute &  Value \\
\hline
System Weight & 25Kg \\
Overall Length & 60cm \\
Overall Width & 42cm \\
Overall Height & 94cm \\
Max Velocity & 20cm/s \\
Payload & 0.5kg\\
Structure Material & Aluminum cast \\
Communication & SPI-UART-I2C-Ethernet \\
Voltage connection & 16.8 Vdc \\
Arm Link Speed & 45 Degree/s \\
\hline
\end{tabular}
\end{table}
\newpage
%
% ---- Bibliography ----
%
% BibTeX users should specify bibliography style 'splncs04'.
% References will then be sorted and formatted in the correct style.
%
% \bibliographystyle{splncs04}
% \bibliography{bibo.bib}

\begin{thebibliography}{}
\bibitem{keselman2017intel}Keselman, L., Woodfill, J., Grunnet-Jepsen, A. \& Bhowmik, A. Intel RealSense Stereoscopic Depth Cameras.  (2017), https://arxiv.org/abs/1705.05548
\bibitem{author2024kinematic}
Taheri, H., Qiao, B. \& Ghaeminezhad, N. Kinematic Model of a Four Mecanum Wheeled Mobile Robot. {\em International Journal Of Computer Applications}. \textbf{113} pp. 6-9 (2015,3)
\bibitem{Kraetzschmar2014}Kraetzschmar, G., Hochgeschwender, N., Nowak, W., Hegger, F., Schneider, S., Dwiputra, R., Berghofer, J. \& Bischoff, R. RoboCup@Work: Competing for the Factory of the Future.  (2015,5)

\bibitem{KohlbrecherMeyerStrykKlingaufFlexibleSlamSystem2011}Kohlbrecher, S., Von Stryk, O., Meyer, J. \& Klingauf, U. A flexible and scalable SLAM system with full 3D motion estimation. {\em 2011 IEEE International Symposium On Safety, Security, And Rescue Robotics (SSRR)}. (2011,11) 
\bibitem{Hart1968}Hart, P., Nilsson, N. \& Raphael, B. A Formal Basis for the Heuristic Determination of Minimum Cost Paths. {\em IEEE Trans. Syst. Sci. Cybern.}. \textbf{4} pp. 100-107 (1968), https://api.semanticscholar.org/CorpusID:206799161
\bibitem{6968200}Goyal, J. \& Nagla, K. A new approach of path planning for mobile robots. {\em Proceedings Of The 2014 International Conference On Advances In Computing, Communications And Informatics, ICACCI 2014}. pp. 863-867 (2014,9)
\bibitem{1453566}Ang, K., Chong, G. \& Li, Y. PID Control System Analysis, Design, and Technology. {\em Control Systems Technology, IEEE Transactions On}. \textbf{13} pp. 559 - 576 (2005,8)

\bibitem{Jocher_Ultralytics_YOLO_2023}Jocher, G., Chaurasia, A. \& Qiu, J. Ultralytics YOLO.  (2023,1), https://github.com/ultralytics/ultralytics
\bibitem{7780460}Redmon, J., Divvala, S., Girshick, R. \& Farhadi, A. You Only Look Once: Unified, Real-Time Object Detection.  (2016,6)
\bibitem{pca}F.R.S., K. LIII. On lines and planes of closest fit to systems of points in space. {\em Philosophical Magazine Series 1}. \textbf{2} pp. 559-572 (1901), https://api.semanticscholar.org/CorpusID:125037489

\bibitem{ros}Quigley, M., Conley, K., Gerkey, B., Faust, J., Foote, T., Leibs, J., Wheeler, R. \& Ng, A. ROS: an open-source Robot Operating System. {\em ICRA Workshop On Open Source Software}. \textbf{3} (2009,1)
\bibitem{Li_2023}Li, Z. Review of PID control design and tuning methods. {\em Journal Of Physics: Conference Series}. \textbf{2649} pp. 012009 (2023,11)

\end{thebibliography}
%
% \bib

\end{document}